\documentclass[letterpaper, 10 pt, conference]{ieeeconf}  

\IEEEoverridecommandlockouts                              

\usepackage{booktabs}
\usepackage{multirow}
\usepackage{graphics} 
\usepackage{epsfig} 
\usepackage{mathptmx} 
\usepackage{times} 
\usepackage{amsmath} 
\usepackage{amssymb}  
\usepackage[cmyk]{xcolor}
\usepackage{listings}

\usepackage{mathtools}
\usepackage{hyperref}
\usepackage{mathrsfs}
\usepackage{bbm}

\usepackage{algorithm}
\usepackage{algorithmic}

\definecolor{codegreen}{rgb}{0,0.6,0}
\definecolor{codegray}{rgb}{0.5,0.5,0.5}
\definecolor{codepink}{RGB}{252, 142, 172}
\definecolor{codepurple}{rgb}{0.58,0,0.82}
\definecolor{backcolour}{RGB}{245,245,245}
\lstdefinestyle{mystyle}{
    backgroundcolor=\color{backcolour},   
    commentstyle=\color{magenta},
    keywordstyle=\color{blue},
    numberstyle=\tiny\color{codegray},
    stringstyle=\color{codepurple},
    basicstyle=\fontfamily{\ttdefault}\footnotesize,
    breakatwhitespace=false,         
    breaklines=true,                 
    keepspaces=true,    
    frame=single,
    numbersep=5pt,                  
    showspaces=false,                
    showstringspaces=false,
    showtabs=false,                  
    tabsize=2,
    classoffset=1, %
    keywordstyle=\color{violet},
    classoffset=1,
}
\title{\LARGE \bf
CorridorVLA: Explicit Spatial Constraints for Generative Action Heads via Sparse Anchors
}

\author{
	{Dachong Li}$^{1}$
	~~~ {Zhuangzhuang Chen}$^{2}$
	~~~ {Jin Zhang}$^{3}$
	~~~ {Jianqiang Li}$^{3*}$\thanks{*Corresponding author. This work is supported in part by the National Natural Science Funds for Distinguished Young Scholar under Grant 62325307, in part by the National Natural Science Foundation Major Scientific Research Instrument Development Project under Grant 62527809, in part by the Shenzhen Key Industry R\&D Program Project under Grant ZDCY20250901102300001, in part by Guangdong Province Key Areas Research and Development Program under Grants 2025B0909020002, in part by Open Project of State Key Laboratory of Synthetical Automation for Process Industries (SAPI-2025-KFKT-11), in part by the National Natural Science Foundation of China under Grants 62373257, 62473264, 62203134, in part by the Natural Science Foundation of Guangdong Province under Grants 2023B1515120038, in part by Shenzhen Science and Technology Innovation Commission (KJZD20230923113801004), in part by the Shenzhen University 2025 Graduate Student Independent Innovation Achievement Cultivation Project, in part by the Intelligent Computing Center of Shenzhen University.}  \\
	\textsuperscript{1} College of Computer Science and Software Engineering, Shenzhen University \\ 
	\textsuperscript{2} Hong Kong University of Science and Technology \\ 
	~~\textsuperscript{3}~ School of Artificial Intelligence, Shenzhen University \\ 
        \texttt{\small{lidachong2023@email.szu.edu.cn,
                        lijq@szu.edu.cn
}}}

\begin{document}

\maketitle
\thispagestyle{empty}
\pagestyle{empty}

\begin{abstract}

Vision--Language--Action (VLA) models often use intermediate representations to connect multimodal inputs with continuous control, yet spatial guidance is often injected implicitly through latent features.
We propose \textbf{CorridorVLA}, which predicts sparse \emph{spatial anchors} as incremental physical changes (e.g., $\Delta$-positions) and uses them to impose an explicit tolerance region in the training objective for action generation.
The anchors define a tolerance corridor that guides a flow-matching action head: trajectories whose implied spatial evolution falls outside the corridor receive corrective gradients, while small trajectory deviations remain tolerated by the training objective. CorridorVLA improves SmolVLA by 4.45 percentage points on LIBERO and improves SmolVLA and GR00T by 12.37 and 7.98 percentage points, respectively, on the more challenging LIBERO-Plus benchmark.
Notably, under the same single-policy 4-in-1 setting, where one policy is jointly trained and evaluated across all task suites, GR00T-Corr achieves an 83.21\% success rate.
These results indicate that action-aligned physical cues can provide direct and interpretable constraints for generative action policies, complementing spatial guidance encoded in visual or latent forms.
The code and released model checkpoints are publicly available at \url{https://github.com/lidc54/corridorVLA} and \url{https://huggingface.co/lidc/CorridorVLA}, respectively.
\end{abstract}

\section{Introduction}

Vision--Language--Action (VLA) models have recently drawn increasing attention as a route toward general-purpose robotic policies that unify perception, language grounding, and control.
Early large-scale systems such as RT-2~\cite{BrohanBrownEtAl2023RT2} and OpenVLA~\cite{kim2024openvla} suggest that scaling multimodal backbones can translate into broader task coverage in robotics.
At the same time, the field has been actively experimenting with different design choices---from diffusion/flow-based action heads that improve continuous control fidelity (e.g., Octo~\cite{Ghosh2024Octo}, pi0~\cite{black2024pi0}, RDT~\cite{liu2024rdt1b}), to richer multimodal structures and training signals (e.g., GR-1/GR-2~\cite{wu2023gr1,cheang2024gr2}, RoboDreamer~\cite{zhou2024robodreamer}, and RL-augmented variants~\cite{li2025simplevla_rl,lu2025vlarl}).
These parallel threads reflect an ongoing evolution of VLA paradigms rather than a settled blueprint~\cite{zhang2025pure_vla_survey}.

Alongside architectural progress, latent representations have become an important interface for organizing multimodal information and supporting embodied decision making~\cite{latent_space_survey}. Meanwhile, the robotics community continues to accumulate data from increasingly diverse platforms and setups.
Differences in embodiments, controllers, camera configurations, and annotation conventions make it natural for datasets to expose heterogeneous state/action parameterizations and task-specific idiosyncrasies.
A recurring theme in VLA design is therefore to introduce intermediate representations that capture task-relevant structure in a more shareable form---goal images, affordance-like cues, reward codes, or other abstractions summarized in recent surveys~\cite{zhong2025survey}.
While such representations do not eliminate heterogeneity, they provide a practical interface for transferring common semantics across robots and tasks.

\begin{figure}[t!]
	\begin{center}
		\includegraphics[width=\columnwidth]{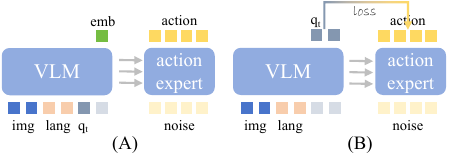}
		\caption{\textbf{Motivation.}
            (\textbf{A}) A common VLA route encodes spatial guidance in an image-style latent: the backbone predicts location-related visual tokens/features that \emph{modulate} the vision--language latent representation, thereby influencing action generation indirectly.
            (\textbf{B}) CorridorVLA explores a lightweight alternative: the backbone predicts sparse key spatial anchors as text-style physical quantities, and these anchors impose an explicit corridor constraint on the downstream action generation objective.}
		\label{motivation}
	\end{center}
\end{figure}

Among candidate intermediates, spatial cues are particularly prominent.
A broad line of work seeks to represent ``what should change'' in the scene---often through future-oriented or change-focused modeling---and use it to support action generation.
For instance, CoTVLA~\cite{zhao2025cot} and DreamVLA~\cite{zhang2025dreamvla} highlight the utility of emphasizing regions of change, and ReconVLA~\cite{song2025reconvla} explores predicting future observations to inform long-horizon behavior.
These approaches encode spatial guidance in visual or latent forms and inject it through representation learning.
Motivated by the same goal of leveraging spatial structure, we explore a complementary route: can spatial guidance be expressed as \emph{direct, text-style} physical quantities that align more closely with the action space, and can such cues constrain action generation at the objective level?
As illustrated in Fig.~\ref{motivation}, unlike visual or latent intermediates that influence action generation implicitly through feature interactions, our formulation predicts sparse action-aligned spatial anchors and uses them to impose an explicit tolerance constraint on the downstream generative action head.

In this paper, we explore this direction through \textbf{CorridorVLA}.
We predict sparse future spatial anchors from the vision-language backbone using learnable slots.
We then use these anchors to impose an explicit tolerance region in the learning objective for action generation: the spatial evolution implied by the generated trajectory is encouraged to stay within the tolerance band, with deviations receiving corrective gradients while small spatial deviations remain tolerated by the training objective.
We instantiate this idea on top of a flow-matching action expert, where the corridor regularizer complements the standard velocity regression objective.

Using SmolVLA~\cite{shukor2025smolvla} as a representative flow-matching policy, we evaluate CorridorVLA on the LIBERO benchmark~\cite{liu2023libero} and observe consistent performance gains over the baseline. These findings indicate that sparse, text-style spatial anchors can serve as effective action-aligned supervision, enabling a direct and interpretable form of spatial guidance for generative action policies.

Our contributions are three-fold:
\begin{itemize}
    \item We propose \textbf{CorridorVLA}, which predicts sparse future spatial anchors as action-aligned physical cues and uses them to constrain action generation through a tolerance-region objective.
    \item We formulate an explicit loss-space coupling between text-style physical cues and a flow-matching action head, complementing prior visual/latent spatial-cue formulations.
    \item We demonstrate consistent improvements across LIBERO and LIBERO-Plus on two VLA backbones, achieving 4.45--12.37 percentage-point gains over their respective base policies, together with ablations that clarify effective design choices.

\end{itemize}


\section{Related Work}
\subsection{Spatially Grounded Intermediate Representations}

Recent progress in Vision--Language--Action (VLA) modeling has been closely tied to how information is represented and organized for embodied decision making. A recent survey from an action-tokenization perspective~\cite{zhong2025survey} summarizes multiple tokenizable forms of multimodal information, reflecting the community effort to build scalable VLA systems under heterogeneous embodiments, sensors, and dataset conventions. In this landscape, a prominent direction is to introduce intermediate representations that help connect high-level multimodal understanding with low-level continuous control.

A considerable body of work uses future-state imagery or video as outputs or intermediate targets, including CoTVLA~\cite{zhao2025cot}, DreamVLA~\cite{zhang2025dreamvla}, and ReconVLA~\cite{song2025reconvla}. These approaches emphasize modeling state transitions and often benefit from the sparsity of predictive signals (e.g., focusing on regions that change). Our work is motivated by a related intuition---spatial evolution provides useful structure---but explores a different instantiation: rather than representing future changes through visual-style intermediates, we study sparse, low-dimensional physical quantities as predictive spatial cues, and further use them to impose an explicit constraint on action generation.

Another line of research strengthens cross-modal reasoning by designing prompts or token layouts that better align vision and language with embodied semantics. For example, InterleaveVLA~\cite{fan2025interleave} interleaves textual and visual tokens to improve cross-modal comprehension. In contrast, we focus less on enriching the input stream and more on shaping a lightweight intermediate signal that is closer to the control space, aiming to provide direct guidance for the downstream action module while keeping the interface compact.

Several recent methods also move representations closer to action generation, either by learning action-oriented latents for downstream policies (e.g., UniVLA~\cite{bu2025univla}) or by formulating policies in purely textual terms (e.g., VLA-0~\cite{goyal2025vla}). 
ReKep~\cite{huang2024rekep} is particularly relevant in its use of language-derived \emph{explicit} spatial constraints, realized as keypoint-based cost functions solved via hierarchical optimization. 
In contrast, CorridorVLA predicts sparse future key positions as physical cues and converts them into a loss-space tolerance corridor that directly guides a generative action head, providing a lightweight and interpretable way to inject spatial objectives into continuous trajectory generation.

Classical constraint-based motion planning methods, such as CHOMP~\cite{chomp}, STOMP~\cite{stomp}, and TrajOpt~\cite{trajopt}, optimize robot trajectories under explicit smoothness, collision-avoidance, or kinematic constraints. In contrast, CorridorVLA does not solve a test-time trajectory optimization problem. 
Instead of optimizing trajectories online, CorridorVLA transfers the idea of explicit spatial constraints into the training objective of a generative VLA policy.

\subsection{View-Centered Spatial Grounding}

Several recent VLA works explore camera-centric or ego-centric formulations that build a unified representation space from the agent’s first-person view, including OC-VLA~\cite{zhang2025grounding}, EgoVLA~\cite{Yang2025EgoVLA}, and cVLA~\cite{argus2025cvla}. 
By treating the camera view as the primary reference frame, these methods aim to align perception with action in a view-consistent manner, which is broadly compatible with our motivation of using grounded representations to connect multimodal inputs and control.

At the same time, camera-centered parameterizations inherit practical variability across platforms: camera resolution, field of view, calibration, and mounting all differ substantially from one robot to another, and the resulting representation space may shift accordingly.
This makes cross-system transfer sensitive to viewpoint and sensor configuration, especially when embodiments differ or the camera undergoes non-negligible motion during execution.
In addition, incorporating motion-related information often requires reasoning about coordinate transforms (e.g., between ego-centric and world frames) and maintaining estimates of pose and extrinsics, which can complicate the pipeline when used as a persistent reference.
Motivated by these considerations, our work instead investigates a compact spatial intermediate expressed as simple physical quantities, aiming to remain interpretable and to couple more directly with the action generator without relying on a camera-defined coordinate system.

\section{Method}

\begin{figure}[t!]
	\begin{center}
		\includegraphics[width=\columnwidth]{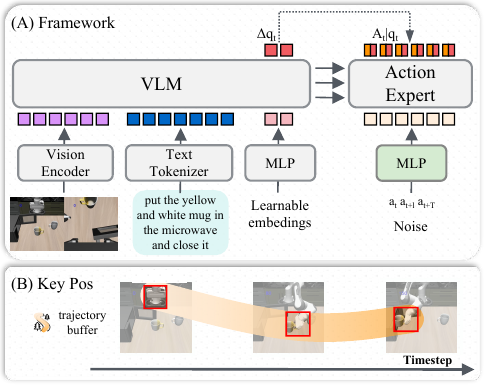}
		\caption{\textbf{Framework.}
            (\textbf{A}) The backbone predicts a small set of future key spatial increments, while the action output is augmented with the corresponding end-effector displacement fields.
            These key increments are then used to constrain action generation, requiring only a few additional prediction slots with minimal changes to the original VLA pipeline.
            (\textbf{B}) Spatial-change guidance provides a simple prior: manipulation trajectories tend to evolve smoothly, so sparse key increments can offer a safe, structured signal that reduces unstructured exploration under stochastic generation.}
		\label{framework}
	\end{center}
\end{figure}

\begin{table}[t]
    \centering
    \caption{
        Main notation used in CorridorVLA.
    }
    \label{tab:notation}
    \resizebox{\columnwidth}{!}{%
        \begin{tabular}{c p{0.73\columnwidth}}
            \toprule
            Symbol & Description \\
            \midrule
            $t$ & Discrete global control-step index in a demonstration trajectory. \\
            $s$ & Continuous flow-matching time, with $s\in[0,1]$. \\
            $\tau$ & Local step index within a length-$T$ action chunk; $\tau_j$ denotes the local index of the $j$-th sparse anchor. \\
            $\mathbf{a}_t$ & Commanded robot action at discrete control step $t$. \\
            $\mathbf{p}_t$ & End-effector Cartesian position at control step $t$. \\
            $\Delta\mathbf{p}_t$ & State-derived EE displacement, defined as $\mathbf{p}_t-\mathbf{p}_{t-1}$. \\
            $\tilde{\Delta\mathbf{p}}$ & Backbone-\emph{predicted} sparse EE displacement anchors for the current action chunk. \\
            $\mathbf{A}$ & Ground-truth extended action chunk. \\
            $\tilde{\mathbf{A}}$ & \emph{Predicted} extended action chunk. \\
            \bottomrule
        \end{tabular}
    }
\end{table}

We view robotic action execution as a structured evolution of spatial states: objects and the end-effector move through a sequence of meaningful configurations before a task is completed. 
Existing VLA systems often provide such guidance through visual or latent intermediates.
These signals can be effective, but they are commonly encoded in visual or latent forms, which may entangle task-relevant motion cues with appearance-level details and typically influence the action head only through implicit feature interactions.
An overview of the proposed CorridorVLA framework is shown in Fig.~\ref{framework}. For clarity, Table~\ref{tab:notation} summarizes the main notation used throughout the method.

In this work, we ask a more direct question: can \emph{text-style} spatial cues, expressed as simple physical quantities of spatial change, serve as an effective intermediate representation for VLA? 
We focus on predicting sparse key waypoints along an execution window and using them as \emph{explicit} spatial constraints during action generation. 
This design aims to (i) keep the intermediate signal close to the control manifold (e.g., incremental displacements rather than images), and (ii) make the guidance act at the objective level, providing a clear training signal beyond latent feature shaping.
To isolate the effect of the proposed representation and objective, we instantiate CorridorVLA on top of SmolVLA~\cite{shukor2025smolvla}. Its lightweight architecture enables fast iteration and fine-grained ablations, while the relatively small model size helps attribute performance gains to the proposed design rather than increased capacity.

Two design principles guide our formulation. First, the spatial cues should be predicted from the same vision--language backbone that conditions the action policy, ensuring that they are grounded in the same multimodal context. Second, rather than serving only as latent features, these cues should directly constrain the action-generation objective by providing explicit trajectory-level supervision.

\subsection{Sparse Key-Position Prediction}
\label{subsec:prediction_setting}

We predict a sparse set of future \emph{spatial anchors} as lightweight physical cues, instantiated as end-effector (EE) 3D $\Delta$-positions at $\mathbf{K}$ selected steps within a length-$\mathbf{T}$ action chunk.
These anchors can in principle be generated autoregressively or predicted through a set of learnable anchor tokens. We adopt the latter: $K$ learnable anchor tokens are appended to the multimodal token sequence, and their contextualized hidden states are mapped to the corresponding $K$ EE anchors. This design predicts all anchors in a single backbone forward pass, avoiding sequential decoding whose inference latency scales with the number of predicted anchors.

We instantiate the EE anchor target as either absolute EE positions or incremental EE position changes.
Absolute positions are tied to a global reference frame and the episode-specific initial EE configuration, making them more demanding to predict. In contrast, incremental displacements describe local motion within the action window and are more directly aligned with the action space.
As shown in Table~\ref{tab:exp1}, predicting EE $\Delta$-positions (\textit{$\Delta$-pos}) consistently outperforms predicting absolute positions (\textit{pos}), and we therefore use $\Delta$-positions as our default anchor representation.

Formally, let $\mathbf{o}_t$ denote the image observation and $\mathbf{l}_t$ the language instruction at step $t$.
We introduce $K$ learnable anchor slots $\mathbf{e}\in\mathbb{R}^{K\times d}$.
The backbone encoder $f_\theta(\cdot)$ takes image, language, and the slots as input, and outputs a fused hidden representation $\mathbf{H}_t\in\mathbb{R}^{N\times d}$ together with predicted sparse EE increments $\tilde{\Delta\mathbf{p}_t}\in\mathbb{R}^{K\times 3}$:
\begin{equation}
\big(\mathbf{H}_t,\ \tilde{\Delta\mathbf{p}_t}\big)
=
f_\theta\!\left(\mathbf{o}_t,\ \mathbf{l}_t,\ \mathbf{e}\right).
\label{eq:delta_pos_pred}
\end{equation}
Here $\tilde{\Delta\mathbf{p}_t}=\{\tilde{\Delta\mathbf{p}}_{t+k}\}_{k=1}^K$ denotes the predicted anchor increments.

Let $\Delta\mathbf{p}_t=\mathbf{p}_t-\mathbf{p}_{t-1}$
denote the ground-truth EE displacement increment in Cartesian space. The $K$ sparse anchor increments are extracted from the dense EE trajectory at indices selected by the Ramer--Douglas--Peucker (RDP) procedure~\cite{douglas1973algorithms}.

We supervise the anchors using
\begin{equation}
\mathcal{L}_{\Delta p}
=
\frac{1}{K}\sum_{k=1}^{K}
\rho\!\left(\left\|\tilde{\Delta\mathbf{p}}-\Delta\mathbf{p}\right\|_2\right),
\label{eq:delta_pos_loss}
\end{equation}
where $\rho(\cdot)$ is a robust penalty (e.g., $\ell_1$ or Huber).

\begin{table}[t]
    \centering
    \caption{
        Success rates (\%) on LIBERO for the 4-in-1 model.
        }
    \label{tab:exp1}
    \resizebox{\columnwidth}{!}{%
        \begin{tabular}{lccccc}
            \toprule
            Method & Long & Goal & Object & Spatial & Avg \\
            \midrule
            SmolVLA-Base & 72.0 & 89.0 & 98.0 & 87.0 & 86.5 \\
            pos &74.6	&90.8	&93.4	&87.2	&86.5 \\ 
            $\Delta$-pos &75.6	&90	&93.6	&90.8	&\textbf{87.5} \\ 
            \bottomrule
        \end{tabular}
}
\end{table}

\subsection{Aligning Action Supervision with Spatial Variability}
\label{subsec:Aligning}

In manipulation, the commanded action $\mathbf{a}_t$ and the realized spatial displacement $\Delta\mathbf{p}_t$ can differ due to actuation biases and intermittent contacts. 
To make supervision better reflect the physical effect of control, we extend the action target with an explicit displacement term. 
Concretely, for each step in an action chunk, we augment the action vector with the corresponding end-effector $\Delta$-position, and denote the resulting extended action as ${\mathbf{ap}}_t \triangleq [\mathbf{a}_t,\Delta\mathbf{p}_t]$. We stack the extended actions over a length-$T$ chunk as ${\mathbf{A}}=[{\mathbf{ap}}_{t+1},\ldots,{\mathbf{ap}}_{t+T}]^\top$.
We refer to this output design as \emph{extra-A}. 
Beyond providing an additional physically grounded training signal, \emph{extra-A} also aligns the action-head supervision with the backbone-predicted sparse anchors in Sec.~\ref{subsec:prediction_setting}, enabling the two components to share a common spatial quantity.

We further combine sparse-anchor prediction with \emph{extra-A} in a merged variant (\emph{merge} in Table~\ref{tab:exp2}). 
Empirically, this combination yields consistent gains, suggesting that explicitly coupling backbone-predicted spatial cues with action-generation supervision is a practical direction for improving generative VLA policies.

\begin{table}[t]
    \centering
    \caption{
        Success rates (\%) on LIBERO for the 4-in-1 model. 
        }
    \label{tab:exp2}
    \resizebox{\columnwidth}{!}{%
        \begin{tabular}{lccccc}
            \toprule
            Method & Long & Goal & Object & Spatial & Avg \\
            \midrule
            SmolVLA-Base & 72.0 & 89.0 & 98.0 & 87.0 & 86.5 \\
            extra-A &76.6	&87	&99.2	&89.8	&88.15 \\
            $\Delta$-pos &75.6	&90	&93.6	&90.8	& 87.5 \\
            merge &79.2	&90.4	&94	&92.4 	&89\\
            \bottomrule
        \end{tabular}
}
\end{table}

\subsection{Flow Matching with Trajectory-Aware Coupling}
\label{subsec:flow_matching}

\begin{figure}[t!]
	\begin{center}
		\includegraphics[width=\columnwidth]{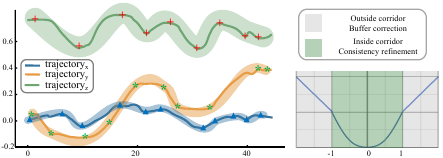}
            \caption{\textbf{Geometric intuition of the proposed tolerance corridor.}
            Sparse anchors define an adaptive corridor around the dense spatial trajectory. Predictions outside the corridor receive buffer correction, while predictions inside the corridor are refined through cumulative-progress consistency.}
		\label{key}
	\end{center}
\end{figure}

Fig.~\ref{key} illustrates the geometric intuition: sparse anchors induce a tolerance corridor around the reference spatial evolution, where predictions outside the corridor receive corrective gradients and predictions inside the corridor are further refined by cumulative-progress consistency.
We train the action expert with flow matching (FM) as in SmolVLA, and couple it with trajectory-level spatial constraints from the same sparse anchors in Sec.~\ref{subsec:prediction_setting} and Sec.~\ref{subsec:Aligning}.
This coupling uses two terms: a corridor \emph{buffer} that defines a tolerant safe region to shrink the stochastic search space, and an in-corridor \emph{consistency} term that continues refining predictions after they enter the buffer.
Together, they behave like a smooth-L1 objective: fast correction outside the corridor and gradual convergence inside.
The overall objective combines the FM loss, the anchor prediction loss (Eq.~\eqref{eq:delta_pos_loss}), and the corridor regularizer.

\paragraph{Flow matching in the extended action space.}

Let $\mathbf{x}=\mathrm{vec}(\mathbf{A})$.
Given Gaussian noise $\boldsymbol{\xi}\sim\mathcal{N}(\mathbf{0},\mathbf{I})$ and $s\sim\mathcal{U}(0,1)$, FM defines
\begin{equation}
\mathbf{z}_s = (1-s)\mathbf{x} + s\,\boldsymbol{\xi}, \qquad s\in[0,1],
\label{eq:fm_interp}
\end{equation}
and learns a time-conditioned velocity field $\mathbf{v}_\theta(\mathbf{z}_s,s)$ via
\begin{equation}
\mathcal{L}_{\mathrm{FM}}
=
\mathbb{E}_{s,\mathbf{x},\boldsymbol{\xi}}
\left[
\left\|
\mathbf{v}_\theta(\mathbf{z}_s,s) - (\boldsymbol{\xi}-\mathbf{x})
\right\|_2^2
\right].
\label{eq:fm_loss}
\end{equation}
Following the standard decoding used in FM action models, we form an estimate of the (vectorized) action sample at time $s$ as
\begin{equation}
\tilde{\mathbf{x}} = \mathbf{z}_s - s\,\mathbf{v}_\theta(\mathbf{z}_s,s),
\qquad
\tilde{\mathbf{A}}=\mathrm{unvec}(\tilde{\mathbf{x}})\in\mathbb{R}^{T\times D}.
\label{eq:xhat_from_v}
\end{equation}

\paragraph{Anchor extraction and corridor buffer.}

Let $\mathbf{A}$ denote the ground-truth extended action chunk corresponding to $\tilde{\mathbf{A}}$, and let $g(\mathbf{A})\in\mathbb{R}^{T\times 3}$ extract the position fields (xyz) from its $T$ steps. We set the corridor width as the scaled maximum point-to-polyline distance from this dense sequence to the polyline formed by the $K$ ground-truth anchors:
\begin{equation}
\delta(\mathbf{A}) = \max_{\tau\in{1,\ldots,T}}
    \operatorname{dist}\left(
    g(\mathbf{A})_{\tau},
    \operatorname{Polyline}\left(
    {\Delta\mathbf{p}_k}
    \right)
    \right)
    \label{eq:delta_def}
\end{equation}
where $\operatorname{dist}(\cdot,\cdot)$ denotes the Euclidean distance from a point to a polyline.

We then penalize violations outside the corridor:
\begin{equation}
\mathcal{L}_{\mathrm{buf}}
=
\frac{1}{K}\sum_{k=1}^{K}
\left[
\delta(\tilde{\mathbf{A}})
- \delta(\mathbf{A})
\right]_+,
\label{eq:buffer_loss}
\end{equation}
where $[\cdot]_+=\max(\cdot,0)$.

\paragraph{In-corridor consistency.}
Once $\tilde{\mathbf{A}}$ enters the corridor, Eq.~\eqref{eq:buffer_loss} becomes inactive.
To keep refining the trajectory and prevent drift within the feasible region, we add a consistency term based on stage-wise cumulative progress.
Let $\mathcal{C}(\cdot)$ denote the cumulative-sum operator applied along time on the same extracted $\Delta$-position sequence, i.e., $\mathcal{C}(g(\mathbf{A}))_\tau=\sum_{j=1}^{\tau} g(\mathbf{A})_j$.
We define
\begin{equation}
\mathcal{L}_{\mathrm{cons}}=
\sum_{j=1}^{K} w_{\tau_j}
\left\|
\mathcal{C}(g(\tilde{\mathbf{A}}))_{\tau_j} -
\mathcal{C}(\Delta\mathbf{p})_{\tau_j}
\right\|_2^2,
\label{eq:cons_loss}
\end{equation}
where $\tau_j\in{1,\ldots,T}$ denotes the trajectory index of the $j$-th sparse anchor, and $w_\tau=\frac{\tau}{\sum_{j=1}^{T}j}=\frac{2\tau}{T(T+1)}$ assigns increasing weight to later anchors.

\paragraph{Noise-aware weighting and overall objective.}
We weight the corridor regularizer by noise level, since geometric constraints are most reliable when the FM state is closer to data.
From Eq.~\eqref{eq:fm_interp}, $\mathbf{z}_s$ becomes increasingly noise-dominated as $s\!\to\!1$, and thus less informative for enforcing spatial consistency.
We therefore use $w(s)=1-s$ to downweight high-noise stages and emphasize the corridor constraints as $s\!\to\!0$.

\begin{equation}
\mathcal{L}_{\mathrm{corr}}= w(s)\big(\mathcal{L}_{\mathrm{buf}}+\mathcal{L}_{\mathrm{cons}}\big).
\label{eq:corrloss}
\end{equation}
The overall training objective is
\begin{equation}
    \mathcal{L}
    =
    \mathcal{L}_{\mathrm{FM}}
    +
    \lambda_{\Delta p}\mathcal{L}_{\Delta p}
    +
    \lambda_{\mathrm{corr}}\mathcal{L}_{\mathrm{corr}}
\label{eq:overall_loss}
\end{equation}
where $\mathcal{L}_{\Delta p}$ is defined in Eq.~\eqref{eq:delta_pos_loss}.

The corridor regularizer is used only during training. At inference time, CorridorVLA follows the same flow-matching sampling procedure as the underlying policy and does not require ground-truth anchors, test-time optimization, or additional trajectory refinement. The predicted anchor slots are produced together with the backbone features, introducing only negligible overhead relative to the base model.

\section{Experiment}

\subsection{Experimental Setup}
\label{subsec:experimental_setup}

We evaluate our method on two representative VLA backbones: \textbf{SmolVLA} and \textbf{GR00T}. 
SmolVLA is implemented using the \textbf{LeRobot} framework~\cite{cadene2024lerobot} (v0.32), while GR00T follows the public implementation provided by StarVLA~\cite{community2026starvla}. 
Unless stated otherwise, we keep the training protocols and hyperparameters identical to the respective official defaults for both backbones, ensuring a fair and reproducible comparison.

Our method introduces a sparse set of future spatial anchors derived from the action chunk. 
Specifically, given the action horizon (chunk size) used by the flow-matching action head, we sample $K$ sparse anchor steps and predict their corresponding spatial increments in the backbone; we use $K{=}3$ by default.
This only requires adding a small number of prediction tokens to the backbone ($K{=}3$ additional tokens in our implementation), while leaving the model capacity and all other settings unchanged.
We conduct experiments on \textbf{LIBERO}~\cite{liu2023libero} and \textbf{LIBERO-Plus}~\cite{fei2025liberoplus}. 
Since the SmolVLA vision encoder operates at $512$ resolution, we re-render LIBERO observations to $512{\times}512$, which allows us to reproduce the reported SmolVLA (0.45B) performance (SR $86.5\%$ vs.\ $87.4\%$ reported). 
For LIBERO-Plus, the released data only supports the default $256{\times}256$ resolution, so all results on LIBERO-Plus are reported under $256$ input resolution.

\subsection{Main Results}
\label{subsec:main_results}
Our method, denoted as \emph{Corr}, mainly modifies the training objective with a corridor-style constraint and leaves the architecture nearly unchanged.
In practice, we add only $K{=}3$ prediction tokens and leave the main architecture unchanged (Table~\ref{tab:libero_results}). 
On LIBERO (Table~\ref{tab:libero_results}), \emph{SmolVLA-Corr} improves success rate by \textbf{4.45} over SmolVLA-Base. No test-time optimization is introduced because the corridor loss is used only during training.

We further test robustness on the more challenging LIBERO-Plus benchmark (Table~\ref{tab:libero_plus_results}).
Since LIBERO-Plus is released at $256{\times}256$ resolution, SmolVLA does not operate under its preferred $512$-resolution setting.
Even so, SmolVLA-Corr achieves a $\textbf{12.37}$ gain over SmolVLA-Base, showing that the corridor constraint remains effective under stronger perturbations and less favorable inputs.
Finally, we validate cross-backbone transfer by applying the same modification to GR00T.
\emph{GR00T-Corr} improves success rate by $\textbf{7.98}$ over GR00T-Base and compares favorably to baselines reported in the LIBERO-Plus benchmark.

\begin{table}[t]
    \centering
    \caption{
        Success rates (\%) on \textbf{LIBERO} for the 4-in-1 model. \emph{Corr} denotes our method.
        }
    \label{tab:libero_results}
    \resizebox{\columnwidth}{!}{%
        \begin{tabular}{lccccc}
            \toprule
            Method & Long & Goal & Object & Spatial & Avg \\
            \midrule
            $\pi$0(3.3B) & 73 & 95.0 & 86.0 & 90.0 & 86.0 \\ 
            GraspVLA~\cite{deng2025graspvla} & 82.0 & 91.2 & 94.1 & 89.1 & 89.1 \\ 
            NORA~\cite{hung2025nora} & 74.6 & 89.4 & 95.4 & 92.2 & 87.9 \\ 
            SmolVLA-Base(0.45B) &  72.0 & 89.0 & 98.0 & 87.0 & 86.5 \\
            \textbf{SmolVLA-Corr}  &85.2	&90.8	&95.8	&92	 & \textbf{90.95} \\
            \bottomrule
        \end{tabular}
}
\end{table}

\begin{table}[t]
    \centering
    \caption{
        Success rates (\%) on \textbf{LIBERO-Plus} for the 4-in-1 model. \emph{Corr} denotes our method. Reported baseline results are reproduced from  the LIBERO-Plus~\cite{fei2025liberoplus} benchmark appendix.
        }
    \label{tab:libero_plus_results}
    \resizebox{\columnwidth}{!}{%
        \begin{tabular}{lccccc}
            \toprule
            Method & Long & Goal & Object & Spatial & Avg \\
            \midrule
            \multicolumn{6}{l}{\textit{SmolVLA comparison group}}\\
            NORA~\cite{hung2025nora}    & 36.3 & 38.8 & 34.4 & 47.6 & 39 \\ 
            UniVLA~\cite{bu2025univla}    & 39.9 & 40.7 & 36.7 & 55.5 & 52.1 \\ 
            SmolVLA-Base & 46.53 & 35.89 & 66.2 & 32.85 & 45.37 \\
            \textbf{SmolVLA-Corr} &49.27	&55.27	&72.36	&54.04	&57.74 \\
            \midrule
            \multicolumn{6}{l}{\textit{GR00T and larger VLA baselines}}\\
            $\pi$0     & 48.4 & 44.9 & 61.4 & 60.7 & 53.6 \\ 
            OpenVLA-OFT~\cite{kim2025finetuningvla}    & 66.4 & 63 & 66.5 & 84 & 69.6 \\ 
            GR00T-Base ~\cite{nvidia2025gr00t} & 62.21 & 68.54 & 84.55 & 85.64 & 75.23 \\
            \textbf{GR00T-Corr} & 74.55	&85.75	&88.4	&84.14	&\textbf{83.21} \\
            \bottomrule
        \end{tabular}
}
\end{table}

\section{Ablation Study}
\label{subsec:ablation}

\subsection{Necessity of Corridor Loss Components}
CorridorVLA augments the standard flow-matching objective with two corridor terms: a buffer constraint and an in-corridor consistency refinement.
A natural question is whether both terms are necessary, or whether the gain mainly comes from one component.
As shown in Table~\ref{tab:ablation}, removing either term causes a clear drop in performance, while using both yields the best results.
The largest improvement from combining both terms appears on the Long suite, suggesting that stable action generation can benefit from both out-of-corridor correction and in-corridor refinement.

By default, we select the $K$ anchor steps using a two-stage simplification: we first apply the Ramer--Douglas--Peucker (RDP) algorithm, a standard polyline simplification method that retains salient points while keeping the trajectory within a prescribed approximation error, and then use a dynamic-programming (DP) minimax selection to down-select exactly $K$ anchors by minimizing the worst-case approximation error along the trajectory. 
In Table~\ref{tab:ablation}, we also evaluate uniform interval sampling, which performs worse, indicating that geometry-aware anchor selection provides more informative supervision than naive spacing.

\begin{table}[t]
    \centering
    \caption{
        Success rate (\%) on \textbf{LIBERO} (4-in-1) with ablated corridor loss components.
        }
    \label{tab:ablation}
    \resizebox{\columnwidth}{!}{%
        \begin{tabular}{lccccc}
            \toprule
            Method & Long & Goal & Object & Spatial & Avg \\
            \midrule
            merge &79.2	&90.4	&94	&92.4 	&89\\
            +$\mathcal{L}_{\mathrm{buf}}$  &80.6	&92.4	&92.6	&92.4	&89.5 \\
            +$\mathcal{L}_{\mathrm{cons}}$ &82.4	&89.2	&97.8	&92.2	&90.4  \\
            +$\mathcal{L}_{\mathrm{buf}}+\mathcal{L}_{\mathrm{cons}} - RDP$ &80.2	&88.2	&95.8	&92.2	&89.1 \\
            +$\mathcal{L}_{\mathrm{buf}}$  +$\mathcal{L}_{\mathrm{cons}}$ &85.2	&90.8	&95.8	&92	 &  90.95 \\
            \bottomrule
        \end{tabular}
}
\end{table}

\subsection{Prediction-as-output and backbone interaction}
\label{subsubsec:ablation_interaction}
To understand how predictive spatial cues should interact with the vision--language backbone, we first replace the state pathway from \emph{encoding-as-input} to \emph{prediction-as-output} (\textit{State-as-Output} in Table~\ref{tab:ablquery}). 
Under the default prefix-style masking used in SmolVLA, state tokens act mainly as suffix conditioning.
Once treated as prediction targets, allowing these predicted tokens to attend bidirectionally to the vision--language context (\textit{State-as-Output+BiAttn}) yields consistent gains.
This suggests that when spatial cues are modeled as prediction targets, richer cross-modal exchange in the backbone can be beneficial, motivating our use of prediction-style anchors with bidirectional interaction.

\subsection{Reference versus prediction burden: what to predict}
\label{subsubsec:ablation_what_to_predict}
We next ask whether ``predicting more'' state information necessarily translates into better guidance. 
Somewhat unexpectedly, jointly predicting both current and future states (\textit{Predict-CF-State}) degrades performance (Table~\ref{tab:ablquery}). 
A plausible explanation is that forecasting high-dimensional states increases the learning burden and can weaken the role of the observed current state as a stable reference, making the auxiliary signal less reliable for downstream action generation.

This motivates a more conservative design: we keep the current state as an input reference and predict only a future cue.
With this setup, the bidirectional variant (\textit{Keep-C/Predict-F (BiAttn)}) 
achieves a higher average success rate than both the causal-masked counterpart (\textit{Keep-C/Predict-F (Causal)}) and the baseline in Table~\ref{tab:ablquery}, indicating that richer cross-modal interaction remains helpful in this setting.

Since retaining the current state restores performance, the difficulty may largely stem from \emph{predicting} an overly complex state representation. 
We therefore probe simpler, action-aligned targets: predicting only the end-effector position (\textit{EE-Pos Anchor}), and further decoupling from absolute offsets by predicting incremental position changes (\textit{EE-$\Delta$Pos Anchor}).
The incremental form achieves the highest average success rate among the tested anchor targets, and we therefore adopt EE $\Delta$-position anchors throughout the paper.

\begin{table}[t]
    \centering
    \caption{
        Success rate (\%) on \textbf{LIBERO} (4-in-1) for ablations of prediction-as-output interaction and anchor targets.
        }
    \label{tab:ablquery}
    \resizebox{\columnwidth}{!}{%
        \begin{tabular}{lccccc}
            \toprule
            Method & Long & Goal & Object & Spatial & Avg \\
            \midrule
            SmolVLA-Base & 72.0 & 89.0 & 98.0 & 87.0 & 86.5 \\
            State-as-Output &69	&89	&95	&88.8	&85.45 \\
            State-as-Output+BiAttn &70.40	&89.60	&94.40	&91	&86.35 \\
            \midrule
            Predict-CF-State &68.4	&88.6	&93.6	&89.8	&\underline{85.1} \\
            Keep-C/Predict-F (Causal) &70.4	&90.8	&95.8	&87.6	&86.15 \\
            Keep-C/Predict-F (BiAttn) &70.8	&90.8	&97.4	&88.2	&\underline{86.8} \\
            EE-Pos Anchor &74.6	&90.8	&93.4	&87.2	&86.5 \\
            EE-$\Delta$Pos Anchor &75.6	&90	&93.6	&90.8	&\textbf{87.5} \\
            \bottomrule
        \end{tabular}
}
\end{table}

\section{Discussion}

Two limitations of this work should be noted.
First, we do not report real-robot experiments.
CorridorVLA is designed as a lightweight modification on top of existing VLA policies---primarily through objective-level constraints and a minimal interface extension---and our study focuses on verifying whether such constraints provide consistent benefits under standard embodied benchmarks.
Real-world deployment, however, depends on additional factors inherited from the base models (e.g., data collection procedures, sim-to-real gaps, and system identification), which are not addressed by a loss-level change alone.
We view real-robot validation as an important next step, particularly to test whether corridor widths and noise-aware weighting should adapt to contact likelihood and uncertainty in physical interaction. Moreover, CorridorVLA is primarily designed for quasi-static manipulation tasks whose end-effector motion can be summarized by a small number of spatial anchors. Extending the framework to highly dynamic behaviors, discontinuous contact events, bimanual manipulation, or mobile manipulation remains an important direction for future work.

Second, we do not provide a head-to-head comparison with spatial-cue designs that rely on image-based or latent visual intermediates, such as InterleaveVLA and ReconVLA.
These methods represent spatial guidance in a different form---often through richer visual signals and heavier generative components---and are typically evaluated under different training budgets and architectural assumptions.
Our goal here is not to replace such approaches, but to probe a complementary question: whether \emph{text-style} spatial cues, expressed as simple physical quantities closer to the action manifold, can directly constrain generative action policies.
The consistent gains we observe across two backbones and two benchmarks suggest that this direction is viable, even with minimal architectural changes.
This points to an alternative design axis for spatial intermediates: beyond shaping hidden features implicitly, spatial objectives can be injected explicitly at the action-generation level through a tolerant corridor that supports fast correction outside the region and gradual refinement within it.

This corridor-based formulation makes spatial guidance explicit and controllable.
Its effectiveness is largely governed by three coupled choices: the anchor representation (we use end-effector $\Delta$-positions as a simple, action-aligned starting point), the corridor schedule that keeps constraints reliable under stochastic FM sampling, and the way gradients are balanced inside versus outside the corridor.
Understanding these factors may provide a practical route to richer, more interpretable intermediate interactions between the vision--language backbone and the action head.

\section{Conclusion}

We presented \textbf{CorridorVLA}, which predicts sparse spatial anchors as action-aligned physical cues and uses them to impose an explicit tolerance constraint for a flow-matching action head.
This objective-level coupling corrects trajectories when their implied spatial evolution violates the tolerance, while allowing moderate trajectory deviations within the learned tolerance corridor.
Across LIBERO and LIBERO-Plus, CorridorVLA improves the corresponding base policies by 4.45--12.37 percentage points. In particular, on LIBERO-Plus, it improves SmolVLA and GR00T by 12.37 and 7.98 percentage points, respectively.

More broadly, our results highlight a complementary design axis for spatial intermediates in VLA: in addition to encoding spatial structure implicitly in visual/latent features, compact physical cues can directly constrain continuous trajectory generation through the training objective.
We hope this perspective encourages further exploration of action-manifold-aligned intermediates for connecting vision--language understanding and robot control.

\bibliographystyle{IEEEtran}
\bibliography{IEEEabrv,root}

@article{zhong2025survey,
  title   = {A Survey on Vision-Language-Action Models: An Action Tokenization Perspective},
  author  = {Zhong, Yifan and Bai, Fengshuo and Cai, Shaofei and Huang, Xuchuan and Chen, Zhang and Zhang, Xiaowei and Wang, Yuanfei and Guo, Shaoyang and Guan, Tianrui and Lui, Ka Nam and others},
  journal = {arXiv preprint arXiv:2507.01925},
  year    = {2025}
}

@inproceedings{zhao2025cot,
  title     = {CoT-VLA: Visual Chain-of-Thought Reasoning for Vision-Language-Action Models},
  author    = {Zhao, Qingqing and Lu, Yao and Kim, Moo Jin and Fu, Zipeng and Zhang, Zhuoyang and Wu, Yecheng and Li, Zhaoshuo and Ma, Qianli and Han, Song and Finn, Chelsea and others},
  booktitle = {Proceedings of the IEEE/CVF Conference on Computer Vision and Pattern Recognition (CVPR)},
  pages     = {1702--1713},
  year      = {2025}
}

@article{zhang2025dreamvla,
  title   = {DreamVLA: A Vision-Language-Action Model Dreamed with Comprehensive World Knowledge},
  author  = {Zhang, Wenyao and Liu, Hongsi and Qi, Zekun and Wang, Yunnan and Yu, Xinqiang and Zhang, Jiazhao and Dong, Runpei and He, Jiawei and Lu, Fan and Wang, He and others},
  journal = {arXiv preprint arXiv:2507.04447},
  year    = {2025}
}

@article{song2025reconvla,
  title   = {ReconVLA: Reconstructive Vision-Language-Action Model as Effective Robot Perceiver},
  author  = {Song, Wenxuan and Zhou, Ziyang and Zhao, Han and Chen, Jiayi and Ding, Pengxiang and Yan, Haodong and Huang, Yuxin and Tang, Feilong and Wang, Donglin and Li, Haoang},
  journal = {arXiv preprint arXiv:2508.10333},
  year    = {2025}
}

@article{fan2025interleave,
  title   = {Interleave-VLA: Enhancing Robot Manipulation with Interleaved Image-Text Instructions},
  author  = {Fan, Cunxin and Jia, Xiaosong and Sun, Yihang and Wang, Yixiao and Wei, Jianglan and Gong, Ziyang and Zhao, Xiangyu and Tomizuka, Masayoshi and Yang, Xue and Yan, Junchi and others},
  journal = {arXiv preprint arXiv:2505.02152},
  year    = {2025}
}

@misc{bu2025univla,
      title={UniVLA: Learning to Act Anywhere with Task-centric Latent Actions}, 
      author={Qingwen Bu and Yanting Yang and Jisong Cai and Shenyuan Gao and Guanghui Ren and Maoqing Yao and Ping Luo and Hongyang Li},
      year={2025},
      eprint={2505.06111},
      archivePrefix={arXiv},
      primaryClass={cs.RO},
      url={https://arxiv.org/abs/2505.06111}, 
}

@article{goyal2025vla,
  title   = {VLA-0: Building State-of-the-Art VLAs with Zero Modification},
  author  = {Goyal, Ankit and Hadfield, Hugo and Yang, Xuning and Blukis, Valts and Ramos, Fabio},
  journal = {arXiv preprint arXiv:2510.13054},
  year    = {2025}
}

@article{zhang2025grounding,
  title   = {Grounding Actions in Camera Space: Observation-Centric Vision-Language-Action Policy},
  author  = {Zhang, Tianyi and Duan, Haonan and Hao, Haoran and Qiao, Yu and Dai, Jifeng and Hou, Zhi},
  journal = {arXiv preprint arXiv:2508.13103},
  year    = {2025}
}

@article{Yang2025EgoVLA,
  title        = {EgoVLA: Learning Vision--Language--Action Models from Egocentric Human Videos},
  author       = {Ruihan Yang and Qinxi Yu and Yecheng Wu and Rui Yan and Borui Li and An-Chieh Cheng and Xueyan Zou and Yunhao Fang and Hongxu Yin and Sifei Liu and Song Han and Yao Lu and Xiaolong Wang},
  journal      = {arXiv preprint arXiv:2507.12440},
  year         = {2025},
  url          = {https://arxiv.org/abs/2507.12440}
}

@article{argus2025cvla,
  title   = {cVLA: Towards Efficient Camera-Space VLAs},
  author  = {Argus, Max and Bratulic, Jelena and Masnavi, Houman and Velikanov, Maxim and Heppert, Nick and Valada, Abhinav and Brox, Thomas},
  journal = {arXiv preprint arXiv:2507.02190},
  year    = {2025}
}

@article{shukor2025smolvla,
  title   = {SmolVLA: A Vision-Language-Action Model for Affordable and Efficient Robotics},
  author  = {Shukor, Mustafa and Aubakirova, Dana and Capuano, Francesco and Kooijmans, Pepijn and Palma, Steven and Zouitine, Adil and Aractingi, Michel and Pascal, Caroline and Russi, Martino and Marafioti, Andres and others},
  journal = {arXiv preprint arXiv:2506.01844},
  year    = {2025}
}

@article{zhang2025pure_vla_survey,
  title={Pure Vision Language Action (VLA) Models: A Comprehensive Survey},
  author={Zhang, Dapeng and Sun, Jin and Hu, Chenghui and Wu, Xiaoyan and Yuan, Zhenlong and Zhou, Rui and Shen, Fei and Zhou, Qingguo},
  journal={arXiv preprint arXiv:2509.19012},
  year={2025}
}

@article{li2025simplevla_rl,
  title   = {SimpleVLA-RL: Scaling Vision-Language-Action (VLA) Training via Reinforcement Learning},
  author  = {Li, Haozhan and Zuo, Yuxin and Yu, Jiale and Zhang, Yuhao and Yang, Zhaohui and Zhang, Kaiyan and Zhu, Xuekai and Zhang, Yuchen and Chen, Tianxing and Cui, Ganqu and Wang, Dehui and Luo, Dingxiang and Fan, Yuchen and Sun, Youbang and Zeng, Jia and Pang, Jiangmiao and Zhang, Shanghang and Wang, Yu and Mu, Yao and Zhou, Bowen and Ding, Ning},
  journal = {arXiv preprint arXiv:2509.09674},
  year    = {2025}
}

@article{lu2025vlarl,
  title   = {VLA-RL: Towards Masterful and General Robotic Manipulation with Scalable Reinforcement Learning},
  author  = {Lu, Guanxing and Chen, Wei and Li, Xinyu and Sun, Zhi and Zhang, Yutong and Yang, Rui and Wang, Shiqi},
  journal = {arXiv preprint arXiv:2505.18719},
  year    = {2025}
}

@article{kim2024openvla,
  title={Openvla: An open-source vision-language-action model},
  author={Kim, Moo Jin and Pertsch, Karl and Karamcheti, Siddharth and Xiao, Ted and Balakrishna, Ashwin and Nair, Suraj and Rafailov, Rafael and Foster, Ethan and Lam, Grace and Sanketi, Pannag and others},
  journal={arXiv preprint arXiv:2406.09246},
  year={2024}
}

@inproceedings{BrohanBrownEtAl2023RT2,
  title     = {RT-2: Vision-Language-Action Models Transfer Web Knowledge to Robotic Control},
  author    = {Anthony Brohan and Noah Brown and Justice Carbajal and Yevgen Chebotar and Xi Chen and Krzysztof Choromanski and Tianli Ding and Danny Driess and Avinava Dubey and Chelsea Finn and Pete Florence and Chuyuan Fu and Montse Gonz{\'a}lez Arenas and Keerthana Gopalakrishnan and Kehang Han and Karol Hausman and Alex Herzog and Jasmine Hsu and Brian Ichter and Alex Irpan and Nikhil Joshi and Ryan Julian and Dmitry Kalashnikov and Yuheng Kuang and Isabel Leal and Lisa Lee and Tsang-Wei Edward Lee and Sergey Levine and Yao Lu and Henryk Michalewski and Igor Mordatch and Karl Pertsch and Kanishka Rao and Krista Reymann and Michael Ryoo and Grecia Salazar and Pannag Sanketi and Pierre Sermanet and Jaspiar Singh and Anikait Singh and Radu Soricut and Huong Tran and Vincent Vanhoucke and Quan Vuong and Ayzaan Wahid and Stefan Welker and Paul Wohlhart and Jialin Wu and Fei Xia and Ted Xiao and Peng Xu and Sichun Xu and Tianhe Yu and Brianna Zitkovich},
  booktitle = {Proceedings of The 7th Conference on Robot Learning (CoRL)},
  series    = {Proceedings of Machine Learning Research},
  volume    = {229},
  pages     = {2165--2183},
  year      = {2023},
  url       = {https://proceedings.mlr.press/v229/zitkovich23a.html},
  note      = {Also available as arXiv:2307.15818}
}

@article{Ghosh2024Octo,
  title = {Octo: An Open-Source Generalist Robot Policy},
  author = {Ghosh, Dibya and Walke, Homer and Pertsch, Karl and Black, Kevin and Mees, Oier and Dasari, Sudeep and Hejna, Joey and Kreiman, Tobias and Xu, Charles and Luo, Jianlan and Tan, You Liang and Chen, Lawrence Yunliang and Sanketi, Pannag and Vuong, Quan and Xiao, Ted and Sadigh, Dorsa and Finn, Chelsea and Levine, Sergey and others},
  journal = {arXiv preprint arXiv:2405.12213},
  year = {2024}
}

@article{black2024pi0,
  title = {$\pi_{0}$: A Vision‐Language‐Action Flow Model for General Robot Control},
  author = {Black, Kevin and Brown, Noah and Driess, Danny and Esmail, Adnan and Equi, Michael and Finn, Chelsea and Fusai, Niccolò and Groom, Lachy and Hausman, Karol and Ichter, Brian and Jakubczak, Szymon and Jones, Tim and Ke, Liyiming and Levine, Sergey and Li-Bell, Adrian and Mothukuri, Mohith and Nair, Suraj and Pertsch, Karl and Shi, Lucy Xiaoyang and Tanner, James and Vuong, Quan and Walling, Anna and Wang, Haohuan and Zhilinsky, Ury},
  journal = {arXiv preprint arXiv:2410.24164},
  year = {2024}
}

@article{liu2024rdt1b,
  title={RDT-1B: a Diffusion Foundation Model for Bimanual Manipulation},
  author={Liu, Songming and Wu, Lingxuan and Li, Bangguo and Tan, Hengkai and Chen, Huayu and Wang, Zhengyi and Xu, Ke and Su, Hang and Zhu, Jun},
  journal={arXiv preprint arXiv:2410.07864},
  year={2024}
}

@article{wu2023gr1,
  title   = {Unleashing Large-Scale Video Generative Pre-training for Visual Robot Manipulation},
  author  = {Wu, Hongtao and Jing, Ya and Cheang, Chilam and Chen, Guangzeng and Xu, Jiafeng and Li, Xinghang and Liu, Minghuan and Li, Hang and Kong, Tao},
  journal = {arXiv preprint arXiv:2312.13139},
  year    = {2023}
}

@article{cheang2024gr2,
  title   = {GR-2: A Generative Video-Language-Action Model with Web-Scale Knowledge for Robot Manipulation},
  author  = {Cheang, Chi-Lam and Chen, Guangzeng and Jing, Ya and Kong, Tao and Li, Hang and Li, Yifeng and Liu, Yuxiao and Wu, Hongtao and Xu, Jiafeng and Yang, Yichu and Zhang, Hanbo and Zhu, Minzhao},
  journal = {arXiv preprint arXiv:2410.06158},
  year    = {2024}
}

@article{zhou2024robodreamer,
  title   = {RoboDreamer: Learning Compositional World Models for Robot Imagination},
  author  = {Zhou, Siyuan and Du, Yilun and Chen, Jiaben and Li, Yandong and Yeung, Dit-Yan and Gan, Chuang},
  journal = {arXiv preprint arXiv:2404.12377},
  year    = {2024}
}

@article{liu2023libero,
  title   = {LIBERO: Benchmarking Knowledge Transfer for Lifelong Robot Learning},
  author  = {Liu, Bo and Zhu, Yifeng and Gao, Chongkai and Feng, Yihao and Liu, Qiang and Zhu, Yuke and Stone, Peter},
  journal = {arXiv preprint arXiv:2306.03310},
  year    = {2023}
}

@article{cadene2024lerobot,
  title   = {LeRobot: State-of-the-art Machine Learning for Real-World Robotics in PyTorch},
  author  = {Cadène, Rémi and Alibert, Simon and Soare, Alexander and Gallouedec, Quentin and Zouitine, Adil and Palma, Steven and Kooijmans, Pepijn and Aractingi, Michel and Shukor, Mustafa and Aubakirova, Dana and Russi, Martino and Capuano, Francesco and Pascal, Caroline and Moss, Jess and Wolf, Thomas},
  journal = {arXiv preprint arXiv:2510.12403},
  year    = {2025}
}

@misc{huang2024rekep,
      title={ReKep: Spatio-Temporal Reasoning of Relational Keypoint Constraints for Robotic Manipulation}, 
      author={Wenlong Huang and Chen Wang and Yunzhu Li and Ruohan Zhang and Li Fei-Fei},
      year={2024},
      eprint={2409.01652},
      archivePrefix={arXiv},
      primaryClass={cs.RO},
      url={https://arxiv.org/abs/2409.01652}, 
}

@misc{fei2025liberoplus,
      title={LIBERO-Plus: In-depth Robustness Analysis of Vision-Language-Action Models}, 
      author={Senyu Fei and Siyin Wang and Junhao Shi and Zihao Dai and Jikun Cai and Pengfang Qian and Li Ji and Xinzhe He and Shiduo Zhang and Zhaoye Fei and Jinlan Fu and Jingjing Gong and Xipeng Qiu},
      year={2025},
      eprint={2510.13626},
      archivePrefix={arXiv},
      primaryClass={cs.RO},
      url={https://arxiv.org/abs/2510.13626}, 
}

@misc{deng2025graspvla,
      title={GraspVLA: a Grasping Foundation Model Pre-trained on Billion-scale Synthetic Action Data}, 
      author={Shengliang Deng and Mi Yan and Songlin Wei and Haixin Ma and Yuxin Yang and Jiayi Chen and Zhiqi Zhang and Taoyu Yang and Xuheng Zhang and Wenhao Zhang and Heming Cui and Zhizheng Zhang and He Wang},
      year={2025},
      eprint={2505.03233},
      archivePrefix={arXiv},
      primaryClass={cs.RO},
      url={https://arxiv.org/abs/2505.03233}, 
}

@misc{hung2025nora,
      title={NORA: A Small Open-Sourced Generalist Vision Language Action Model for Embodied Tasks}, 
      author={Chia-Yu Hung and Qi Sun and Pengfei Hong and Amir Zadeh and Chuan Li and U-Xuan Tan and Navonil Majumder and Soujanya Poria},
      year={2025},
      eprint={2504.19854},
      archivePrefix={arXiv},
      primaryClass={cs.RO},
      url={https://arxiv.org/abs/2504.19854}, 
}

@misc{kim2025finetuningvla,
      title={Fine-Tuning Vision-Language-Action Models: Optimizing Speed and Success}, 
      author={Moo Jin Kim and Chelsea Finn and Percy Liang},
      year={2025},
      eprint={2502.19645},
      archivePrefix={arXiv},
      primaryClass={cs.RO},
      url={https://arxiv.org/abs/2502.19645}, 
}

@misc{nvidia2025gr00t,
      title={GR00T N1: An Open Foundation Model for Generalist Humanoid Robots}, 
      author={NVIDIA and : and Johan Bjorck and Fernando Castañeda and Nikita Cherniadev and Xingye Da and Runyu Ding and Linxi "Jim" Fan and Yu Fang and Dieter Fox and Fengyuan Hu and Spencer Huang and Joel Jang and Zhenyu Jiang and Jan Kautz and Kaushil Kundalia and Lawrence Lao and Zhiqi Li and Zongyu Lin and Kevin Lin and Guilin Liu and Edith Llontop and Loic Magne and Ajay Mandlekar and Avnish Narayan and Soroush Nasiriany and Scott Reed and You Liang Tan and Guanzhi Wang and Zu Wang and Jing Wang and Qi Wang and Jiannan Xiang and Yuqi Xie and Yinzhen Xu and Zhenjia Xu and Seonghyeon Ye and Zhiding Yu and Ao Zhang and Hao Zhang and Yizhou Zhao and Ruijie Zheng and Yuke Zhu},
      year={2025},
      eprint={2503.14734},
      archivePrefix={arXiv},
      primaryClass={cs.RO},
      url={https://arxiv.org/abs/2503.14734}, 
}

@article{community2026starvla,
  title={StarVLA: A Lego-like Codebase for Vision-Language-Action Model Developing},
  author={Community, StarVLA},
  journal={arXiv preprint arXiv:2604.05014},
  year={2026},
  eprint={2604.05014},
  archivePrefix={arXiv},
  primaryClass={cs.RO}
}

@article{chomp,
author = {Zucker, Matt and Ratliff, Nathan and Dragan, Anca D. and Pivtoraiko, Mihail and Klingensmith, Matthew and Dellin, Christopher M. and Bagnell, J. Andrew and Srinivasa, Siddhartha S.},
title = {CHOMP: Covariant Hamiltonian optimization for motion planning},
year = {2013},
issue_date = {August-September 2013},
publisher = {Sage Publications, Inc.},
address = {USA},
volume = {32},
number = {9–10},
issn = {0278-3649},
url = {https://doi.org/10.1177/0278364913488805},
doi = {10.1177/0278364913488805},
journal = {Int. J. Rob. Res.},
month = aug,
pages = {1164–1193},
numpages = {30}
}

@INPROCEEDINGS{stomp,
  author={Kalakrishnan, Mrinal and Chitta, Sachin and Theodorou, Evangelos and Pastor, Peter and Schaal, Stefan},
  booktitle={2011 IEEE International Conference on Robotics and Automation}, 
  title={STOMP: Stochastic trajectory optimization for motion planning}, 
  year={2011},
  volume={},
  number={},
  pages={4569-4574},
  doi={10.1109/ICRA.2011.5980280}}

@article{trajopt,
author = {Schulman, John and Duan, Yan and Ho, Jonathan and Lee, Alex and Awwal, Ibrahim and Bradlow, Henry and Pan, Jia and Patil, Sachin and Goldberg, Ken and Abbeel, Pieter},
title = {Motion planning with sequential convex optimization and convex collision checking},
year = {2014},
issue_date = {August    2014},
publisher = {Sage Publications, Inc.},
address = {USA},
volume = {33},
number = {9},
issn = {0278-3649},
url = {https://doi.org/10.1177/0278364914528132},
doi = {10.1177/0278364914528132},
journal = {Int. J. Rob. Res.},
month = aug,
pages = {1251–1270},
numpages = {20}
}

@misc{latent_space_survey,
      title={The Latent Space: Foundation, Evolution, Mechanism, Ability, and Outlook}, 
      author={Xinlei Yu and Zhangquan Chen and Yongbo He and Tianyu Fu and Guanting Dong and Cheng Yang and Chengming Xu and Yue Ma and Xiaobin Hu and Zhe Cao and Jie Xu and Guibin Zhang and Jiale Tao and Jiayi Zhang and Siyuan Ma and Kaituo Feng and Haojie Huang and Youxing Li and Ronghao Chen and Huacan Wang and Chenglin Wu and Zikun Su and Xiaogang Xu and Kelu Yao and Kun Wang and Chen Gao and Yue Liao and Ruqi Huang and Tao Jin and Zhucun Xue and Cheng Tan and Jiangning Zhang and Wenqi Ren and Yanwei Fu and Yong Liu and Yu Wang and Xiangyu Yue and Yu-Gang Jiang and Shuicheng Yan},
      year={2026},
      eprint={2604.02029},
      archivePrefix={arXiv},
      primaryClass={cs.AI},
      url={https://arxiv.org/abs/2604.02029}, 
}

@inbook{douglas1973algorithms,
    author = {Douglas, David H. and Peucker, Thomas K.},
    publisher = {John Wiley \& Sons, Ltd},
    isbn = {9780470669488},
    title = {Algorithms for the Reduction of the Number of Points Required to Represent a Digitized Line or its Caricature},
    booktitle = {Classics in Cartography},
    chapter = {2},
    pages = {15-28},
    doi = {https://doi.org/10.1002/9780470669488.ch2},
    url = {https://onlinelibrary.wiley.com/doi/abs/10.1002/9780470669488.ch2},
    eprint = {https://onlinelibrary.wiley.com/doi/pdf/10.1002/9780470669488.ch2},
    year = {2011}
}

\end{document}